\documentclass[10pt,twocolumn,letterpaper, arxiv]{article}

\usepackage[arxiv]{cvpr}      

\usepackage{graphicx}
\usepackage{amsmath}
\usepackage{amssymb}
\usepackage{booktabs}

\usepackage[pagebackref,breaklinks,colorlinks]{hyperref}

\usepackage[capitalize]{cleveref}
\crefname{section}{Sec.}{Secs.}
\Crefname{section}{Section}{Sections}
\Crefname{table}{Table}{Tables}
\crefname{table}{Tab.}{Tabs.}

\def\cvprPaperID{6} 
\def\confName{CVPR}
\def\confYear{2023}

\begin{document}

\title{Detecting Images Generated  by Diffusers}

\author{Davide Alessandro Coccomini, Andrea Esuli, Fabrizio Falchi, Claudio Gennaro, Giuseppe Amato\\\\
ISTI-CNR, Pisa, Italy \\\\
\tt\small davidealessandro.coccomini@isti.cnr.it, andrea.esuli@isti.cnr.it,\\ 
\tt\small fabrizio.falchi@isti.cnr.it, claudio.gennaro@isti.cnr.it, giuseppe.amato@isti.cnr.it}
\maketitle

\begin{abstract}
This paper explores the task of detecting images generated by text-to-image diffusion models. To evaluate this, we consider images generated from captions in the MSCOCO and Wikimedia datasets using two state-of-the-art models: Stable Diffusion and GLIDE. Our experiments show that it is possible to detect the generated images using simple Multi-Layer Perceptrons (MLPs), starting from features extracted by CLIP, or traditional Convolutional Neural Networks (CNNs). We also observe that models trained on images generated by Stable Diffusion can detect images generated by GLIDE relatively well, however, the reverse is not true. Lastly, we find that incorporating the associated textual information with the images rarely leads to significant improvement in detection results but that the type of subject depicted in the image can have a significant impact on performance. This work provides insights into the feasibility of detecting generated images, and has implications for security and privacy concerns in real-world applications. The code to reproduce our results is available at: https://github.com/davide-coccomini/Detecting-Images-Generated-by-Diffusers
\end{abstract}

\section{Introduction}
\label{sec:intro}
The ability to generate synthetic images has advanced significantly in recent years, with the development of various techniques such as Generative Adversarial Networks (GANs) and Diffusion Models. These techniques have made it possible to generate high-quality images that are indistinguishable from real ones. However, with this advancement comes the concern about the potential misuse of synthetic images for malicious purposes, such as deepfakes and spreading misinformation. In order to mitigate these concerns, it is crucial to develop robust techniques for detecting synthetic images. The ability to distinguish synthetic images from real ones is essential for maintaining the integrity of information and for protecting individuals from the malicious use of synthetic media. 
The explosion of recent text-to-image methods and their easy access to the general public is leading society towards a point where a good deal of online content is synthetic and where the line between reality and fiction will no longer be so clear. 
In this paper, we present a first attempt to distinguish between generated and real images on the basis of the image itself and the text associated with it and used to describe and generate it. We also analyze the peculiarities of image and text that can lead to a more or less credible image that is difficult to identify.

\begin{figure*}[t]
    \centering
    \includegraphics[width=1\linewidth]{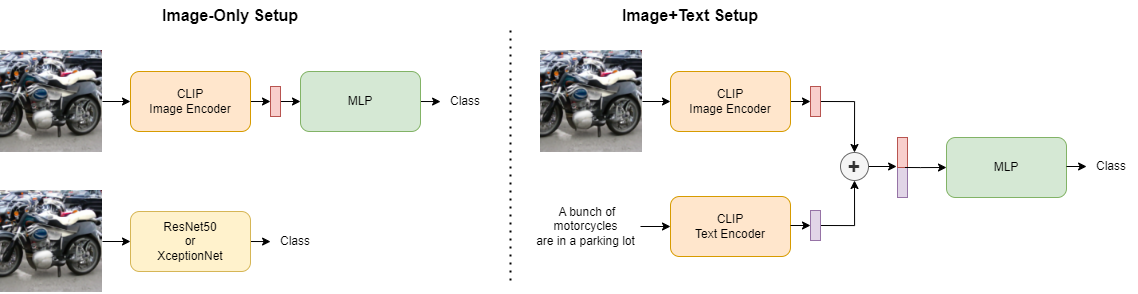} 
    \caption{The two different network setups are shown in the figure. On the left is the Image-Only Setup in which only the image under consideration is used as input to the network while ignoring its associated caption. This setup is used for some MLPs and for the two convolutional networks under consideration. On the right, on the other hand, the Image+Text Setup is presented in which the features obtained from the image are concatenated with those obtained from the caption, both extracted via CLIP and given as input to the network. The latter setup is used exclusively in the training of some MLPs.}
    \label{fig:training}
\end{figure*}

\section{Related Works}
\label{sec:relatedworks}
\subsection{Images Generation}
Synthetic image generation has been a rapidly growing area of research in computer vision and machine learning, with numerous recent advances in deep learning-based generative models such as Generative Adversarial Networks (GANs). The goal of these models is to generate synthetic images that are indistinguishable from real images, with applications in areas such as data augmentation, style transfer, and semantic manipulation.

One of the earliest GAN-based methods for synthetic image generation was introduced by Goodfellow et al. \cite{goodfellow2014generative}. Their model consisted of a generator network that synthesized images, and a discriminator network that learned to distinguish between real and synthetic images. They have been used for several tasks like face synthesis\cite{ruiz2020morphgan, mokhayeri2019crossdomain}, style transfer \cite{xu2021drbgan} and super resolution \cite{ledig2017photorealistic}. Impressive results have been achieved in \cite{karras2020analyzing} where the authors proposed style-based GAN architecture (StyleGAN), capable of generating more credible images. Several variations of GAN architecture have been proposed over the years, for example, the CycleGAN proposed in \cite{Zhu_2017_ICCV}. This architecture is able to perform the task of image-to-image translation optimizing a cycle loss and also allowing the revert of the mapping process.
GAN architecture has been also used for the task of text-to-image translation in which a text describing a context is converted into an image. In particular, in \cite{mirrorgan} the MirrorGAN is proposed, in which the authors exploit the concept of re-description in the sense that the image generated starting from a text, should be describable by another text which is similar to the source one. So the model needs to learn to generate images whose re-description matches as much as possible with the requested one.
Recently a novel architecture was introduced to perform similar tasks, namely Diffusion Models. Diffusion models generate images by refining an initial noise vector through multiple diffusion steps. For text-to-image generation, a given text is encoded into a latent vector used as the initial noise. Some of these models obtained unprecedent results such as DALL-E \cite{pmlr-v139-ramesh21a}, GLIDE \cite{nichol2022glide} and Stable Diffusion \cite{Rombach_2022_CVPR}. These models allow unprecedented control of image generation allowing the users to create very credible images with a high level of detail.

\subsection{Syntethic Images Detection}
The growing credibility and diffusion of generated images raised some concerns in the research community which tried to develop methods capable of effectively distinguishing synthetic content from pristine one. For many time the main concerns were focused around deepfakes with a lot of works proposed trying to detect them \cite{mintime, ftcn, crossforgery, jimaging8100263, Baxevanakis2022TheMD, 9156865, coccomini2022combining, CALDELLI202131, 10.1145/3556384.3556416, Cozzolino2020IDRevealID}. Otherwise, with the recent advancement of generation techniques, attention has been posed also on generic generated content without a fundamental focus on images depicting humans. For example, in \cite{sha2022fake} the authors tried to detect images generated by some diffusion models in two setups, image-only and image+text, where text is the description used to generate the image, being capable of effectively distinguishing between real and generated images. In \cite{corvi2022detection} the researchers tried to train some binary classifiers to distinguish images generated by diffusion models and GAN models. The results highlighted how it is pretty feasible to detect images when the generated method is used in the training set while there is a huge generalization problem. Indeed, the classifiers seem to learn some kind of trace specific for each generator and so are pretty limited when tested on images generated with other methods.

\section{Experiments}
In this section, we explain all the experiments conducted and the details to reproduce them. 
\subsection{Classifiers}
To carry out our experiments, we selected some simple Deep Learning architectures to use as binary classifiers (real or generated image). In particular, a first model used is a simple MLP that takes as input the features extracted from the image by CLIP \cite{radford2021learning} encoders. CLIP (Contrastive Language-Image Pre-Training) is a neural network model that learns to associate natural language descriptions with images. CLIP can use both ResNet50 and Vision Transformer (ViT) features to represent the images it processes and has been choose as a model to extract the features for our classifier because it is able to produce very expressive features both for text and images. The second category of trained models are standard Convolutional Neural Networks widely used in Computer Vision i.e. Resnet50 and XceptionNet both pretrained for image classification on ImageNet datasets.
All models considered have a similar number of parameters to allow for a rigorous comparison.

\begin{table*}
\centering
\begin{tabular}{c|c|c|c|c|c|c|c}
\hline
\hline
$\textbf{Model}$ & $\textbf{Dataset}$ & $\textbf{Mode}$ & $\textbf{Features}$ & $\textbf{Accuracy} \uparrow$ & $\textbf{AUC} \uparrow$ & $\textbf{Params}$ & $\textbf{Pretrain}$ \\
\hline
MLP-Base & MSCOCO & Image Only & CLIP-VIT & 79.5 & 88.8 & 23M & N/A \\
MLP-Base & MSCOCO & Text+Image & CLIP-VIT & 78.5 & 88.8 & 23M & N/A \\
MLP-Base & MSCOCO & Image Only & CLIP-R50 & 67.5 & 75.0 & 23M & N/A \\
MLP-Base & MSCOCO & Text+Image & CLIP-R50 & 66.5 & 74.2 & 23M & N/A \\
XceptionNet & MSCOCO & Image Only & XceptionNet & 94.6 & 98.9 & 20M & ImageNet \\
Resnet50 & MSCOCO & Image Only & Resnet50 & \textbf{97.1} & \textbf{99.6} & 23M & ImageNet \\
& & & & & & & \\[-0.7em]
\hline
& & & & & & & \\[-0.7em]
MLP-Base & Wikipedia & Image Only & CLIP-VIT & 72.8 & 81.4 & 23M & N/A \\
MLP-Base & Wikipedia & Text+Image & CLIP-VIT & 73.1 & 80.8 & 23M & N/A \\
MLP-Base & Wikipedia & Image Only & CLIP-R50 & 65.9 & 74.2 & 23M & N/A \\
MLP-Base & Wikipedia & Text+Image & CLIP-R50 & 64.5 & 73.5 & 23M & N/A \\
XceptionNet & Wikipedia & Image Only & XceptionNet & 90.7 & 97.1 & 20M & ImageNet \\
Resnet50 & Wikipedia & Image Only & Resnet50 & \textbf{94.5} & \textbf{98.1} & 23M & ImageNet \\
\hline
\hline

\end{tabular}
\caption{Results on the test set of the various classifiers trained and tested on real images and images generated with Stable Diffusion. For each row, the dataset considered, training mode used, features extracted via CLIP or convolutional layers, accuracy and AUC, number of model parameters, and whether pretraining was used are indicated.}
\label{tab:stable_diffusion_results}
\end{table*}

\begin{table*}
\centering
\begin{tabular}{c|c|c|c|c|c|c|c}
\hline
\hline

$\textbf{Model}$ & $\textbf{Dataset}$ & $\textbf{Mode}$ & $\textbf{Features}$ & $\textbf{Accuracy} \uparrow$ & $\textbf{AUC} \uparrow$ & $\textbf{Params}$ & $\textbf{Pretrain}$ \\
\hline
MLP-Base & MSCOCO & Image Only & CLIP-VIT & 95.8 & 99.2 & 23M & N/A \\
MLP-Base & MSCOCO & Text+Image & CLIP-VIT & 95.8 & 99.2 & 23M & N/A \\
MLP-Base & MSCOCO & Image Only & CLIP-R50 & 79.0 & 87.4 & 23M & N/A \\
MLP-Base & MSCOCO & Text+Image & CLIP-R50 & 78.0 & 86.4 & 23M & N/A \\
XceptionNet & MSCOCO & Image Only & XceptionNet & 98.9 & 99.9 & 20M & ImageNet \\
Resnet50 & MSCOCO & Image Only & Resnet50 & \textbf{99.3} & \textbf{99.9} & 23M & ImageNet \\
& & & & & & & \\[-0.7em]
\hline 
& & & & & & & \\[-0.7em]
MLP-Base & Wikipedia & Image Only & CLIP-VIT & 93.7 & 98.4 & 23M & N/A \\
MLP-Base & Wikipedia & Text+Image & CLIP-VIT & 94.3 & 98.4 & 23M & N/A \\
MLP-Base & Wikipedia & Image Only & CLIP-R50 & 77.1 & 85.2 & 23M & N/A \\
MLP-Base & Wikipedia & Text+Image & CLIP-R50 & 75.5 & 84.5 & 23M & N/A \\
XceptionNet & Wikipedia & Image Only & XceptionNet & 99.2 & 99.9 & 20M & ImageNet \\
Resnet50 & Wikipedia & Image Only & Resnet50 & \textbf{99.5} & \textbf{99.9} & 23M & ImageNet \\
\hline
\hline

\end{tabular}
\caption{Results on the test set of the various classifiers trained and tested on real images and images generated with GLIDE. For each row, the dataset considered, training mode used, features extracted via CLIP or convolutional layers, accuracy and AUC, number of model parameters, and whether pretraining was used are indicated.}
\label{tab:glide_results}
\end{table*}

\subsection{Dataset}
\begin{figure}[t]
    \centering
    \includegraphics[width=1\linewidth]{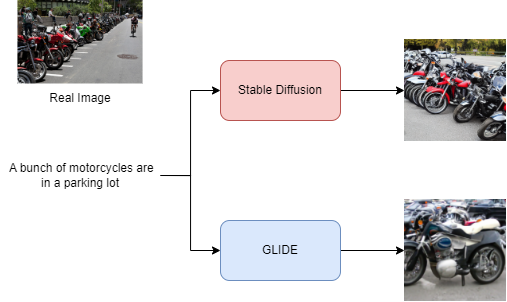} 
    \caption{An example of a caption associated with an image from the MSCOCO dataset given as input to Stable Diffusion and GLIDE to generate two additional images is shown in the figure.}
    \label{fig:generation}
\end{figure}
To validate the ability of a classifier to effectively identify images generated from text, we considered two starting datasets, namely MSCOCO \cite{lin2014microsoft} composed by over 330000 English captioned images and Wikimedia Image-Caption Matching dataset\footnote{\url{https://www.kaggle.com/c/wikipedia-image-caption/overview}} based on Wikipedia Image Text Dataset (WIT) \cite{srinivasan2021wit} and contains 37.6 million entity-rich image-text examples with 11.5
million unique images across 108 Wikipedia languages. Both of these datasets consist of images from many different contexts and each of them is associated with a caption describing it as shown in Figure \ref{fig:generation}. 
For each of the two datasets, 6000 images were extracted from the training set and a further 6000 images were generated using two text-to-image methods, namely Stable Diffusion and GLIDE. The same was done with 1500 images from the validation set and another 6000 from the test. Therefore, all models were trained on a training set consisting of 12000 images, half of which were generated by Stable Diffusion and the same have been done with another training set with images generated by GLIDE. To perform a more detailed analysis of the behaviour of the classifiers considered, the images from the Wikimedia dataset and the corresponding generated images were categorised on the basis of the Wikipedia ontologies available online.  

\subsection{Training Setup}
The classifiers considered were simple Multi Layer Perceptrons (MLPs) or widely used convolutional networks, in particular ResNet50 and XceptionNet. The former are trained from scratch while the convolutional networks are pre-trained on ImageNet. All models are trained with a learning rate of 0.1 decreasing to 0.001, for up to 270 epochs on an NVIDIA A100.
MLPs are trained in two possible setups, image-only and image+text. For MLPs, CLIP-extracted features are exploited, in the first case from the image-only, and in the second case from both the image and the associated caption. In the image+text case, the features are concatenated before being given as input to the model as shown in Figure \ref{fig:training}, which is then able to see both the textual and visual components in a single vector.
Instead, the CNNs considered are trained exclusively in image-only mode and the features used for classification are those resulting from the convolutional layers of the considered architecture.

\section{Results}
In this section we show the results obtained in two main contexts, intra-method and cross-method. 

\subsection{Classification}
In Table \ref{tab:stable_diffusion_results} are shown the performances, in terms of accuracy and AUC, of the classifiers considered when trained and tested with real images and images generated by Stable Diffusion, according to the setup illustrated above. As can be seen from the results, the pretrained models exhibit remarkable capabilities in identifying generated images by acting as perfect detectors.
On the other hand, MLPs achieve satisfactory results especially when CLIP features extracted via a Vision Transformer are used, without the use of pretrain on large datasets. 
Another element that greatly influences the result, especially for MLPs is certainly the dataset, in fact, images from Wikimedia seem to be more difficult to distinguish. This probably results from the high variety of images and contexts in this dataset, however, even in this case the models manage to achieve good levels of accuracy.

\begin{figure}[t]
    \centering
    \includegraphics[width=0.8\linewidth]{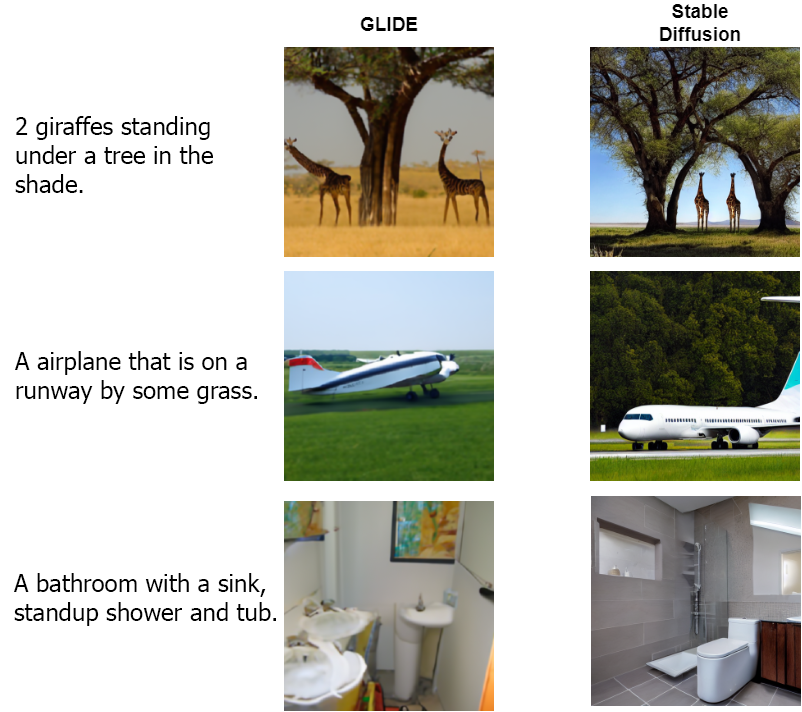}
    \caption{A comparison between the images generated by GLIDE and those via Stable Diffusion from the same caption.}
    \label{fig:comparison}
\end{figure}

As shown instead in Table \ref{tab:glide_results}, the images generated via GLIDE seem to be much easier to detect on both datasets considered and also without the use of pretraining. 
In fact, the MLP achieves an accuracy of 95.8\%, not far from the 99.5\% of pretrained convolutional networks. As can be seen in Figure \ref{fig:comparison} from the same caption, the images generated by GLIDE appear to be more artefactual and bogus than those generated with Stable Diffusion, which obtains more credible results. Just as the latter are more difficult to identify with the naked eye, the models experience similar difficulties.

\begin{table*}[t]
\centering
\begin{tabular}{c|c|c|c|c|c|c}
\hline
\hline

$\textbf{Model}$ & $\textbf{Training Method}$ & $\textbf{Testing Method}$ & $\textbf{Mode}$ & $\textbf{Features}$ & $\textbf{Accuracy} \uparrow$ & $\textbf{AUC} \uparrow$ \\
\hline
MLP-Base & Stable Diffusion & GLIDE & Image-Only & CLIP-R50 & 50.8 & 50.3\\
MLP-Base & Stable Diffusion & GLIDE & Image+Text & CLIP-R50 & 49.9 & 50.9\\
MLP-Base & Stable Diffusion & GLIDE & Image-Only & CLIP-VIT & 64.9 & 75.7\\
MLP-Base & Stable Diffusion & GLIDE & Image+Text & CLIP-VIT & \textbf{69.7} & 76.2\\
XceptionNet & Stable Diffusion & GLIDE & Image-Only & XceptionNet & 58.9 & 76.1\\
Resnet50 & Stable Diffusion & GLIDE & Image-Only & Resnet50 & 61.4 & \textbf{86.1}\\
& & & & & & \\[-0.7em]
\hline
& & & & & & \\[-0.7em]
MLP-Base & GLIDE & Stable Diffusion & Image-Only & CLIP-R50 & 51.5 & 51.7\\
MLP-Base & GLIDE & Stable Diffusion & Image+Text & CLIP-R50 & 50.6 & 51.1\\
MLP-Base & GLIDE & Stable Diffusion & Image-Only & CLIP-VIT & 48.8 & 45.9\\
MLP-Base & GLIDE & Stable Diffusion & Image+Text & CLIP-VIT & \textbf{52.3} & 61.5\\
XceptionNet & GLIDE & Stable Diffusion & Image-Only & XceptionNet & 50.2 & \textbf{73.5}\\
Resnet50 & GLIDE & Stable Diffusion & Image-Only & Resnet50 & 50.2 & 64.6\\
\hline
\hline
\end{tabular}
\caption{Results obtained by the considered classifiers trained on real images and generated with Stable Diffusion and then tested on real images and generated with GLIDE, and vice versa on the MSCOCO dataset.}
\label{tab:cross}
\end{table*}

\begin{table*}[t]
\centering
\begin{tabular}{c|c|c|c|c|c|c}
\hline
\hline

$\textbf{Model}$ & $\textbf{Training Method}$ & $\textbf{Testing Method}$ & $\textbf{Mode}$ & $\textbf{Features}$ & $\textbf{Accuracy} \uparrow$ & $\textbf{AUC} \uparrow$ \\
\hline
MLP-Base   & Stable Diffusion  &      GLIDE       &  Image-Only  & CLIP-R50   & 36.4    & 44.8 \\
MLP-Base   & Stable Diffusion  &      GLIDE       &  Image+Text  & CLIP-R50   & 36.7    & 45.7 \\
MLP-Base   & Stable Diffusion  &      GLIDE       &  Image-Only  & CLIP-VIT   & 46.2    & 40.5 \\
MLP-Base   & Stable Diffusion  &      GLIDE       &  Image+Text  & CLIP-VIT   & 43.4    & 34.4 \\
XceptionNet & Stable Diffusion  &      GLIDE       &  Image-Only & XceptionNet & 48.9    & 48.8 \\
Resnet50   & Stable Diffusion  &      GLIDE       &  Image-Only  & Resnet50  & \textbf{50.9} & \textbf{55.0} \\
           &                   &                  &              &            &          &       \\[-0.7em]
           \hline
           
& & & & & & \\[-0.7em]
MLP-Base   & GLIDE            & Stable Diffusion  &  Image-Only  & CLIP-R50   &  45.6   & 39.9 \\
MLP-Base   & GLIDE            & Stable Diffusion  &  Image+Text  & CLIP-R50   &  39.4   & 45.8 \\
MLP-Base   & GLIDE            & Stable Diffusion  &  Image-Only  & CLIP-VIT   &  48.8   & 45.9 \\
MLP-Base   & GLIDE            & Stable Diffusion  &  Image+Text  & CLIP-VIT   &  48.0   & 45.8 \\
XceptionNet & GLIDE           & Stable Diffusion  &  Image-Only & XceptionNet & \textbf{49.3}  & 49.6 \\
Resnet50   & GLIDE            & Stable Diffusion  &  Image-Only  & Resnet50   &  49.2  &  \textbf{53.7} \\
\hline
\hline

\end{tabular}
\caption{Results obtained by the considered classifiers trained on real images and generated with Stable Diffusion and then tested on real images and generated with GLIDE, and vice versa on the Wikimedia dataset.}
\label{tab:cross_2}
\end{table*}

\begin{table*}[h]
\centering
\begin{tabular}{c|c|c|c|c|c|c|c|c|c|c}
\hline
\hline

\textbf{Model} & \textbf{Mode} & \textbf{Features} & \multicolumn{4}{c|}{\textbf{Stable Diffusion}} & \multicolumn{4}{c}{\textbf{GLIDE}} \\
& & & \multicolumn{2}{c|}{Animated} & \multicolumn{2}{c|}{Inanimate} & \multicolumn{2}{c|}{Animated} & \multicolumn{2}{c}{Inanimate} \\
& & & $FN\downarrow$ & $FP\downarrow$ & $FN\downarrow$ & $FP\downarrow$ & $FN\downarrow$ & $FP\downarrow$ & $FN\downarrow$ & $FP\downarrow$ \\
\hline
MLP-Base & Image-Only & CLIP-R50 & 26.4 & 8.3 & 31.0 & 16.4 & 15.9 & 8.3 & 20.6 & 16.1 \\
MLP-Base & Text+Image & CLIP-R50 & 28.6 & 7.6 & 35.0 & 15.7 & 17.6 & 7.5 & 23.3 & 15.7 \\
MLP-Base & Image-Only & CLIP-VIT & 21.7 & 14.3 & 25.0 & 16.6 & 5.6 & 3.0 & 7.0 & 5.0 \\
MLP-Base & Text+Image & CLIP-VIT & 17.6 & 19.4 & 21.0 & 18.7 & 2.6 & 6.6 & 3.8 & 7.6 \\
XceptionNet & Image-Only & XceptionNet & 6.0 & 7.9 & 7.4 & 7.0 & 1.1 & 0.1 & 2.8 & 1.0 \\
Resnet50 & Image-Only & Resnet50 & 6.0 & 3.0 & 5.8 & 4.0 & 0.2 & 0.2 & 1.1 & 1.0 \\
\hline
\hline

\end{tabular}
\caption{Percentage false negatives and false positive in the two categories considered, animate and inanimate objects, on Wikimedia test dataset with images generated with Stable Diffusion or GLIDE.}
\label{fig:categories_errors}
\end{table*}

\subsection{Cross-Method Classification}
The trained models have shown great ability to identify images generated by text-to-image systems, however, in this section, we want to find out whether they are able to generalize the concept of a generated image to such an extent that they can identify images created with different generators. Indeed, many text-to-image systems are available, and a real-world deployed model should be able to identify generated images regardless of the method used. The danger is in fact that these classifiers learn to distinguish some sort of trace, noise or imprint left by the generator instead of focusing on more general, high-level anomalies or inconsistencies, thus rendering them useless in the real world.
To valid this, we tried training models on real images and images generated with Stable Diffusion and tested them on images generated with GLIDE, and vice versa.

The results reported in Table \ref{tab:cross} show a fair ability on the part of MLPs as well as convolutional networks to identify images generated by GLIDE despite being trained with images generated by Stable Diffusion, on the MSCOCO dataset. As the images generated by the latter are probably more challenging than those obtained with GLIDE, the models are able to exploit the knowledge gained in the training context. Two models stand out in terms of accuracy and AUC, namely the MLP trained in the image+text setup and the Resnet50. The former has a substantial positive difference from the same model trained in the image-only setup. This leads us to believe that the introduction of the text component may have played a role in helping the model to better detect the generated images. 
The opposite experiment, however, led to unsatisfactory results; in fact, the same models trained with GLIDE-generated images achieve very poor accuracy values. The high AUC value, however, suggests that the models trained in this way have not learned the concept of generated images well and end up giving a large number of false positives with a consequent lowering of accuracy.

The same experiments were conducted on Wikimedia datasets as illustrated in Table \ref{tab:cross_2}. In this context, the difficulties that the models had already encountered in the previous experiments become even more evident and stems probably in the greater challenge of the dataset. As it is much more varied than MSCOCO, it makes generalization more difficult, as the models have to learn to distinguish between images generated from very varied contexts. In fact, practically no model manages to exceed the accuracy of 50\%, demonstrating a total inability to generalize. 
\begin{table}[t]
\resizebox{\columnwidth}{!}{%
\begin{tabular}{c|c}
\hline
\hline
\textbf{Feature} & \textbf{Description} \\
\hline
LENGTH & The length of the caption \\
ADJ & Number of adjectives (e.g. big, old, green) \\
ADP & Number of prepositions (e.g. in, to, during) \\
ADV & Number of adverbs (e.g. very, tomorrow, down) \\
AUX & Number of auxiliaries (e.g. is, will do, should do) \\
CCONJ & Number of coordinating conjunctions (e.g. and, or, but)\\
DET & Number of determiner (e.g. a, an, the)\\
INTJ & Number of interjections (e.g. psst, ouch, hello) \\
NOUN & Number of nouns (e.g. girl, cat, air) \\
NUM & Number of numbers (e.g. 1995, 7, seventy, XXII) \\
PART & Number of particles (e.g. 's, not) \\
PRON & Number of pronouns (e.g. I, she, you) \\
PROPN & Number of proper nouns (e.g. James, USA, NATO) \\
PUNCT & Number of punctuations (e.g. ., ?, !) \\
SCONJ & Number of subordinating conjunction (e.g. if, while, that) \\
SYM & Number of symbols (e.g. @, \$, -)\\
VERB & Number of verbs (e.g. runs, fly, eat) \\
X & Number of other type of constructs \\
SPACE  & Number of spaces \\
STOPS & Number of stop words (e.g. and, by, of) \\
NON\_ALPHA & Number of non alphabetic words (e.g. 1, 1997, 33) \\
NAMED\_ENTITIES & Number of named entities  (London, Gary, UE)\\
\hline
\hline

\end{tabular}
}
\caption{Features used for the linguistic analysis conducted with their acronyms.}
\label{tab:variables}
\end{table}

\begin{figure}[t]
    \centering
    \includegraphics[width=1\linewidth]{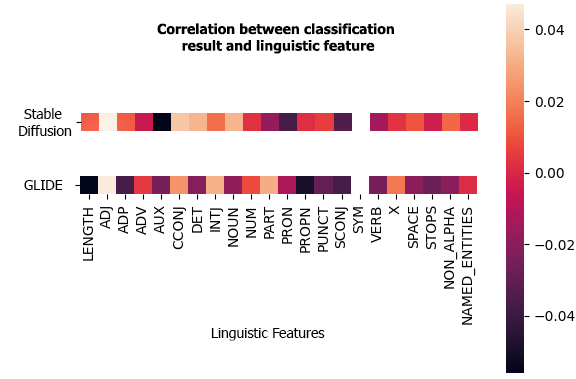}
    \caption{Pearson correlation values between the linguistic features under consideration and the classification provided by an MLP trained in the image+text setup are shown in the figure. The first row refers to the dataset composed of real images and images generated by Stable Diffusion while the second row refers to the dataset with images generated by GLIDE.}
    \label{fig:correlations}
\end{figure}

\subsection{Error Analysis}
The structure of the Wikimedia dataset allows us to perform an error analysis by category. As mentioned above, we have predictively categorized the images in this dataset into various categories (e.g. Artist, City, Road, River, Animal etc) based on the ontologies provided by Wikipedia and the tags associated with each image. They were then further grouped into two macro-categories namely inanimate and animate objects. Based on these categories, we analyzed the errors committed by the classifiers on the test set.

In Figure \ref{fig:categories_errors} we give an example of the percentage false negative, namely undetected generated images, and percentage false positive, namely wrongly classified real images, made on the Wikimedia test set of trained models. Looking at the false negative results, the models tend to have more difficulty to identify generated images when the image depicts an inanimate object (e.g. buildings, roads, objects, infrastructure, rivers etc) than when it is an image with animate subjects (people, animals etc). In other words, then images generated by both models depicting inanimate subjects are more believable and therefore more difficult to distinguish from real ones. Generating, for example, a credible person or animal in the eyes of the classifier thus seems to be a more difficult task for the text-to-image systems under consideration. On the other hand, an inanimate context can have so much variety and anomalies or inconsistencies introduced by generators could easily be mistaken for normality or peculiarities of the scene.

\subsection{Linguistic Analysis}
To better understand the nature of the errors made by the classifiers, we investigated linguistically the captioning associated with them in an attempt to understand whether the patterns are influenced by particular elements in the sentences. The correlation between the correct or incorrect classification of an image and the linguistic features listed and described in Table \ref{tab:variables} was analyzed.

Figure \ref{fig:correlations} shows examples of Pearson correlation between the classification of an MLP trained on real images and images generated by one of the two methods analyzed (Stable Diffusion and GLIDE) and the linguistic features considered. To do so, these were extracted from each caption associated with the test images and the correlation between their variation and the classification provided by the model was derived. As can be seen from the figure, there is no strong correlation between these features and the classification. It thus appears to be more influenced by the category of the image rather than the composition of the sentence.

\section{Conclusions}
In this study, we conducted some analysis on the detection of content generated by text-to-image systems, particularly Stable Diffusion and GLIDE. We tested several classifiers between MLPs and Convolutional Neural Networks highlighting how classical deep learning models are easily able to distinguish images generated with these systems when they have seen examples of them in the training set. By screening their generalization ability, however, they are rarely able to identify images generated by methods other than those used to construct the training set thus highlighting an important issue for these systems to be adopted in the real world.  An analysis of the correlation between the credibility of the generated images and the category to which they belong as well as the composition of the caption associated with them was also conducted. From our experiments, the images generated by both generators considered are more credible when they depict inanimate objects and thus result in greater error on the part of the classifiers. Conversely, images depicting people, animals or animate subjects in general are easier to identify. In addition, there does not appear to be a strong correlation between the linguistic composition of the sentence and the classification ability of the models. Although CLIP is the most natural model to use in a multimodal context such as ours, in order to better investigate the impact of language properties on classification, one possibility would be to use a language model such as RoBERTa to extract features from the captions in the image+text setup. Models such as this might be able to better capture linguistic features, thus translating into a greater impact in the mixed setup classification. We will conduct this and other tests in future works.

\section*{Acknowledgments}
This work was partially supported by project SERICS (PE00000014) under
the NRRP MUR program funded by the EU - NGEU and by FAIR (PE00000013)
funded by the European Commission under the NextGenerationEU. It was also partially supported by AI4Media project, funded by the EC (H2020 - Contract n. 951911).

{\small
\bibliographystyle{ieee_fullname}
\bibliography{egbib}

\begin{thebibliography}{10}\itemsep=-1pt

\bibitem{Baxevanakis2022TheMD}
Spyridon Baxevanakis, Giorgos Kordopatis-Zilos, Panagiotis Galopoulos, Lazaros
  Apostolidis, Killian Levacher, I.~Baris Schlicht, Denis Teyssou, Ioannis
  Kompatsiaris, and Symeon Papadopoulos.
\newblock The mever deepfake detection service: Lessons learnt from developing
  and deploying in the wild.
\newblock In {\em Proceedings of the 1st International Workshop on Multimedia
  AI against Disinformation}, 2022.

\bibitem{CALDELLI202131}
Roberto Caldelli, Leonardo Galteri, Irene Amerini, and Alberto {Del Bimbo}.
\newblock Optical flow based cnn for detection of unlearnt deepfake
  manipulations.
\newblock {\em Pattern Recognition Letters}, 146:31--37, 2021.

\bibitem{coccomini2022combining}
Davide Coccomini, Nicola Messina, Claudio Gennaro, and Fabrizio Falchi.
\newblock Combining efficientnet and vision transformers for video deepfake
  detection, 2022.

\bibitem{crossforgery}
Davide~Alessandro Coccomini, Roberto Caldelli, Fabrizio Falchi, Claudio
  Gennaro, and Giuseppe Amato.
\newblock Cross-forgery analysis of vision transformers and cnns for deepfake
  image detection.
\newblock In {\em Proceedings of the 1st International Workshop on Multimedia
  AI against Disinformation}, MAD '22, page 52–58, New York, NY, USA, 2022.
  Association for Computing Machinery.

\bibitem{mintime}
Davide~Alessandro Coccomini, Giorgos~Kordopatis Zilos, Giuseppe Amato, Roberto
  Caldelli, Fabrizio Falchi, Symeon Papadopoulos, and Claudio Gennaro.
\newblock Mintime: Multi-identity size-invariant video deepfake detection.
\newblock arXiv, 2022.

\bibitem{corvi2022detection}
Riccardo Corvi, Davide Cozzolino, Giada Zingarini, Giovanni Poggi, Koki Nagano,
  and Luisa Verdoliva.
\newblock On the detection of synthetic images generated by diffusion models,
  2022.

\bibitem{Cozzolino2020IDRevealID}
Davide Cozzolino, Andreas Rossler, Justus Thies, Matthias Niessner, and Luisa
  Verdoliva.
\newblock Id-reveal: Identity-aware deepfake video detection.
\newblock {\em 2021 IEEE/CVF International Conference on Computer Vision
  (ICCV)}, pages 15088--15097, 2020.

\bibitem{goodfellow2014generative}
Ian Goodfellow, Jean Pouget-Abadie, Mehdi Mirza, Bing Xu, David Warde-Farley,
  Sherjil Ozair, Aaron Courville, and Yoshua Bengio.
\newblock Generative adversarial nets.
\newblock {\em Advances in neural information processing systems}, pages
  2672--2680, 2014.

\bibitem{jimaging8100263}
Luca Guarnera, Oliver Giudice, Francesco Guarnera, Alessandro Ortis, Giovanni
  Puglisi, Antonino Paratore, Linh M.~Q. Bui, Marco Fontani, Davide~Alessandro
  Coccomini, Roberto Caldelli, Fabrizio Falchi, Claudio Gennaro, Nicola
  Messina, Giuseppe Amato, Gianpaolo Perelli, Sara Concas, Carlo Cuccu, Giulia
  Orrù, Gian~Luca Marcialis, and Sebastiano Battiato.
\newblock The face deepfake detection challenge.
\newblock {\em Journal of Imaging}, 8(10), 2022.

\bibitem{karras2020analyzing}
Tero Karras, Samuli Laine, Miika Aittala, Janne Hellsten, Jaakko Lehtinen, and
  Timo Aila.
\newblock Analyzing and improving the image quality of stylegan, 2020.

\bibitem{ledig2017photorealistic}
Christian Ledig, Lucas Theis, Ferenc Huszar, Jose Caballero, Andrew Cunningham,
  Alejandro Acosta, Andrew Aitken, Alykhan Tejani, Johannes Totz, Zehan Wang,
  and Wenzhe Shi.
\newblock Photo-realistic single image super-resolution using a generative
  adversarial network, 2017.

\bibitem{9156865}
Lingzhi Li, Jianmin Bao, Hao Yang, Dong Chen, and Fang Wen.
\newblock Advancing high fidelity identity swapping for forgery detection.
\newblock In {\em 2020 IEEE/CVF Conference on Computer Vision and Pattern
  Recognition (CVPR)}, pages 5073--5082, 2020.

\bibitem{lin2014microsoft}
Tsung-Yi Lin, Michael Maire, Serge Belongie, James Hays, Pietro Perona, Deva
  Ramanan, Piotr Doll{\'a}r, and C~Lawrence Zitnick.
\newblock Microsoft coco: Common objects in context.
\newblock {\em European conference on computer vision}, pages 740--755, 2014.

\bibitem{mokhayeri2019crossdomain}
Fania Mokhayeri, Kaveh Kamali, and Eric Granger.
\newblock Cross-domain face synthesis using a controllable gan, 2019.

\bibitem{nichol2022glide}
Alex Nichol, Prafulla Dhariwal, Aditya Ramesh, Pranav Shyam, Pamela Mishkin,
  Bob McGrew, Ilya Sutskever, and Mark Chen.
\newblock Glide: Towards photorealistic image generation and editing with
  text-guided diffusion models, 2022.

\bibitem{mirrorgan}
Tingting Qiao, Jing Zhang, Duanqing Xu, and Dacheng Tao.
\newblock Mirrorgan: Learning text-to-image generation by redescription.
\newblock In {\em 2019 IEEE/CVF Conference on Computer Vision and Pattern
  Recognition (CVPR)}, pages 1505--1514, 2019.

\bibitem{radford2021learning}
Alec Radford, Jong~Wook Kim, Chris Hallacy, Aditya Ramesh, Gabriel Goh,
  Sandhini Agarwal, Girish Sastry, Amanda Askell, Pamela Mishkin, Jack Clark,
  Gretchen Krueger, and Ilya Sutskever.
\newblock Learning transferable visual models from natural language
  supervision, 2021.

\bibitem{pmlr-v139-ramesh21a}
Aditya Ramesh, Mikhail Pavlov, Gabriel Goh, Scott Gray, Chelsea Voss, Alec
  Radford, Mark Chen, and Ilya Sutskever.
\newblock Zero-shot text-to-image generation.
\newblock In Marina Meila and Tong Zhang, editors, {\em Proceedings of the 38th
  International Conference on Machine Learning}, volume 139 of {\em Proceedings
  of Machine Learning Research}, pages 8821--8831. PMLR, 18--24 Jul 2021.

\bibitem{Rombach_2022_CVPR}
Robin Rombach, Andreas Blattmann, Dominik Lorenz, Patrick Esser, and Bj\"orn
  Ommer.
\newblock High-resolution image synthesis with latent diffusion models.
\newblock In {\em Proceedings of the IEEE/CVF Conference on Computer Vision and
  Pattern Recognition (CVPR)}, pages 10684--10695, June 2022.

\bibitem{ruiz2020morphgan}
Nataniel Ruiz, Barry-John Theobald, Anurag Ranjan, Ahmed~Hussein Abdelaziz, and
  Nicholas Apostoloff.
\newblock Morphgan: One-shot face synthesis gan for detecting recognition bias,
  2020.

\bibitem{sha2022fake}
Zeyang Sha, Zheng Li, Ning Yu, and Yang Zhang.
\newblock De-fake: Detection and attribution of fake images generated by
  text-to-image diffusion models.
\newblock {\em arXiv preprint arXiv:2210.06998}, 2022.

\bibitem{srinivasan2021wit}
Krishna Srinivasan, Karthik Raman, Jiecao Chen, Michael Bendersky, and Marc
  Najork.
\newblock Wit: Wikipedia-based image text dataset for multimodal multilingual
  machine learning.
\newblock In {\em Proceedings of the 44th International ACM SIGIR Conference on
  Research and Development in Information Retrieval}, pages 2443--2449, 2021.

\bibitem{xu2021drbgan}
Wenju Xu, Chengjiang Long, Ruisheng Wang, and Guanghui Wang.
\newblock Drb-gan: A dynamic resblock generative adversarial network for
  artistic style transfer, 2021.

\bibitem{10.1145/3556384.3556416}
Ying Zhang, Linyu Guan, Jingyan Lu, and Songyang Wu.
\newblock Face forgery detection of deepfake based on multiple convolutional
  neural networks.
\newblock SPML '22, page 211–217, New York, NY, USA, 2022. Association for
  Computing Machinery.

\bibitem{ftcn}
Yinglin Zheng, Jianmin Bao, Dong Chen, Ming Zeng, and Fang Wen.
\newblock Exploring temporal coherence for more general video face forgery
  detection.
\newblock In {\em ICCV}, pages 15024--15034, 2021.

\bibitem{Zhu_2017_ICCV}
Jun-Yan Zhu, Taesung Park, Phillip Isola, and Alexei~A. Efros.
\newblock Unpaired image-to-image translation using cycle-consistent
  adversarial networks.
\newblock In {\em Proceedings of the IEEE International Conference on Computer
  Vision (ICCV)}, Oct 2017.

\end{thebibliography}
}

\end{document}